\documentclass{article}
\usepackage{ijcai23}

\usepackage{times}
\usepackage{soul}
\usepackage{url}
\usepackage[hidelinks]{hyperref}
\usepackage[utf8]{inputenc}
\usepackage[small]{caption}
\usepackage{graphicx}
\usepackage{amsmath}
\usepackage{amsthm}
\usepackage{booktabs}
\usepackage{algorithm}
\usepackage{algorithmic}
\usepackage[switch]{lineno}
\usepackage{graphicx}
\usepackage{latexsym}
\usepackage{times}
\usepackage{soul}
\usepackage[table]{xcolor}
\usepackage{url}
\usepackage[hidelinks]{hyperref}
\usepackage[utf8]{inputenc}
\usepackage{caption}
\DeclareCaptionFont{small}{\small} 
\usepackage{graphicx}
\usepackage{amsmath}
\usepackage{amsthm}
\usepackage{booktabs}
\usepackage{algorithm}
\usepackage{algorithmic}
\usepackage{times}
\usepackage{soul}
\usepackage{url}
\usepackage[hidelinks]{hyperref}
\usepackage[utf8]{inputenc}
\usepackage{graphicx}
\usepackage{amsmath}
\usepackage{amssymb}
\usepackage{booktabs}
\usepackage{algorithm}
\usepackage{algorithmic}
\usepackage[switch]{lineno}
\usepackage{color}
\usepackage{array}
\usepackage{booktabs}
\usepackage{multirow}
\usepackage{bm}
\usepackage{mathrsfs}
\usepackage{enumitem}
\usepackage{amsthm}

\title{Alleviating Attention Hacking in Discriminative Reward Modeling through Interaction Distillation}

\author{
Jianxiang Zang
\affiliations
College of Computer Science and Artificial Intelligence, Fudan University\\
\emails
\url{jxzang25@m.fudan.edu.cn}
}

\begin{document}

\maketitle

\begin{abstract}
The reward model (RM), as the core component of reinforcement learning from human feedback (RLHF) for large language models (LLMs), responsible for providing reward signals to generated responses. However, the mainstream discriminative reward modeling is inadequate in terms of token-level interaction, making its judgment signals vulnerable to being hacked by misallocated attention to context. This stems from two fundamental limitations: (1) Current preference modeling employs decoder-only architectures, where the unidirectional causal attention mechanism leads to forward-decaying intra-sequence attention within the prompt-response sequence. (2) The independent Siamese-encoding paradigm induces the absence of token-level inter-sequence attention between chosen and rejected sequences.
To address this “attention hacking”, we propose “\textit{\textbf{Interaction Distillation}}”, a novel training framework for more adequate discriminative reward modeling via attention-level optimization. The method introduces an interaction-based natural language understanding model as the teacher to provide sophisticated token interaction patterns via comprehensive attention, and guides the reward modeling to simulate teacher model's interaction pattern through an attentional alignment objective. Through extensive experiments, interaction distillation has demonstrated its ability to provide more stable and generalizable reward signals compared to state-of-the-art RM optimization methods that target data noise, highlighting the attention hacking constitute a more fundamental limitation in discriminative RM.

\end{abstract}


\section{Introduction}

Recently, reinforcement learning from human feedback (RLHF) has been widely adopted to enhance the alignment between large language model (LLM) behaviors and human preferences~\cite{ouyang2022training,shen2023large,xu2025survey}. The reward model (RM), as the core component of RLHF, is responsible for providing reward signals for LLM-generated responses~\cite{wang2024secrets,dou2024multi,zhong2025comprehensive}, guiding LLMs to produce responses with higher reward scores. The mainstream discriminative RM training first concatenates the prompt and response into a sequence, uses a LLM as the backbone to obtain sequence representations, and maps them to a scalar reward score via a linear layer. It then enforces the Bradley-Terry-based preference modeling~\cite{bradley1952rank} constraint to ensure that the chosen response receives a higher reward score than the rejected one~\cite{schulman2017proximal,wang2024secrets}. However, this preference modeling is inadequate at the token interaction level, as both intra-sequence and inter-sequence attention remain incomplete during training. This incompleteness makes the reward signals vulnerable to being hacked by misallocated attention to context. 

This “\textbf{attention hacking}” stems from two interaction limitations: (1) From intra-sequence attention perspective, existing discriminative RMs rely entirely on Transformer decoders as their backbone, where sequence attention is implemented via \textbf{unidirectional causal multi-head attention}. This forward-decaying attention mechanism prevents the RM from capturing interactions with preceding tokens in the sequence, resulting in disproportionate focus on the latter tokens of the prompt-response sequence. (2) From inter-sequence attention perspective, the \textbf{independent Siamese-encoding paradigm} in preference modeling leads to the absence of early interaction attention between chosen and rejected sequences. This design flaw prevents the RM from capturing token-level interactions, resulting in inadequate cross-sequence semantic perception.


These token-level interaction limitations led us directly to consider the \textbf{interaction-based natural language understanding (NLU) model}, which is based on fine-tuned encoder-only Transformer (the BERT family~\cite{devlin2019bert,he2022deberta}) on NLU datasets~\cite{camburu2018snli}. It performs downstream classification tasks by jointly encoding the concatenation of two sequences. Since it processes the concatenated sequence, its \textbf{bidirectional and global attention} models the most comprehensive interactions between intra-sequence and inter-sequence tokens. Therefore, although interaction-based NLU models with the BERT family as their backbone are significantly smaller in scale compared to RMs based on LLMs, they serve as stronger teacher models in terms of attention. 

Inspired by this and knowledge distillation~\cite{li2022virt,lu2022ernie}, we propose “\textit{\textbf{Interaction Distillation}}”, a novel framework aiming to achieve more adequate token-level interaction in discriminative reward modeling, and the reward model trained with this method is termed \textbf{\textsc{Id-Rm}}. This framework consists of three steps: (1) Introducing an interaction-based NLU model as the teacher model, which unified encodes concatenated sequence pairs. Since it operates on concatenated sequences, its MHA can fully model both intra-sequence and inter-sequence token interactions. Consequently, we capture its global attention map as the learning objective for RM. (2) Leveraging the query and key matrices from preference modeling to simulate the teacher model's comprehensive attention computation, guiding the standard RM to simulate the sophisticated interaction patterns of the teacher model. (3) Designing an attentional optimization objective to align the simulated attention maps from preference modeling with the real and comprehensive attention maps from the interaction-based NLU model, thereby bridging the gap in token-level interaction between RM and the teacher model.

The current state-of-the-art (SOTA) discriminative reward modeling optimization methods primarily address data noise \cite{touvron2023llama,wang2024reward,rame2024warm,miao2024inform}. In contrast, \textsc{Id-Rm} delivers more stable and generalizable reward signals without introducing inference costs, with its advantages manifested in two key aspects: (1) For RLHF tasks \cite{volske2017tl,bai2022training}, the \textsc{Id-Rm}-optimized policy models generate responses that best align with human preferences. (2) In out-of-distribution preference perception tasks \cite{lambert2024rewardbench,zhou2024rmb}, \textsc{Id-Rm} achieves the highest average accuracy. These results highlight that the intrinsic deficiencies in RM's attention mechanisms constitute a more fundamental limiting factor than data noise. In conclusion, our main contributions are summarized as follows:

\begin{itemize}
    \item We systematically analyze the attention hacking in reward modeling, which stems from inadequate token-level interactions: (1) forward-decaying intra-sequence attention caused by unidirectional causal attention; (2) absent token-level inter-sequence attention due to the independent Siamese-encoding paradigm.
    \item We propose interaction distillation, a novel preference modeling method. This method introduces an interaction-based NLU model as the teacher to provide a comprehensive attention map, and guides the preference modeling to simulate this sophisticated interaction pattern through an attentional distillation objective.
    \item Through extensive experiments on both RLHF and OOD preference perception tasks, interaction distillation demonstrates its ability to deliver more stable and generalizable reward signals compared to other SOTA reward modeling optimization approaches, without introducing inference costs.
\end{itemize}


\section{Attention Hacking in Discriminative RM}


In this section, we first introduce the training objective for discriminative reward models. We then analyze how both intra- and inter-sequence attention in preference modeling are directly linked to the reward model's decision. However, the inherent nature of preference modeling is inadequate at the token interaction level, making reward signals vulnerable to being hacked by misallocated attention to context.




\begin{figure}[t]
  \centering
  \includegraphics[width=1\linewidth]{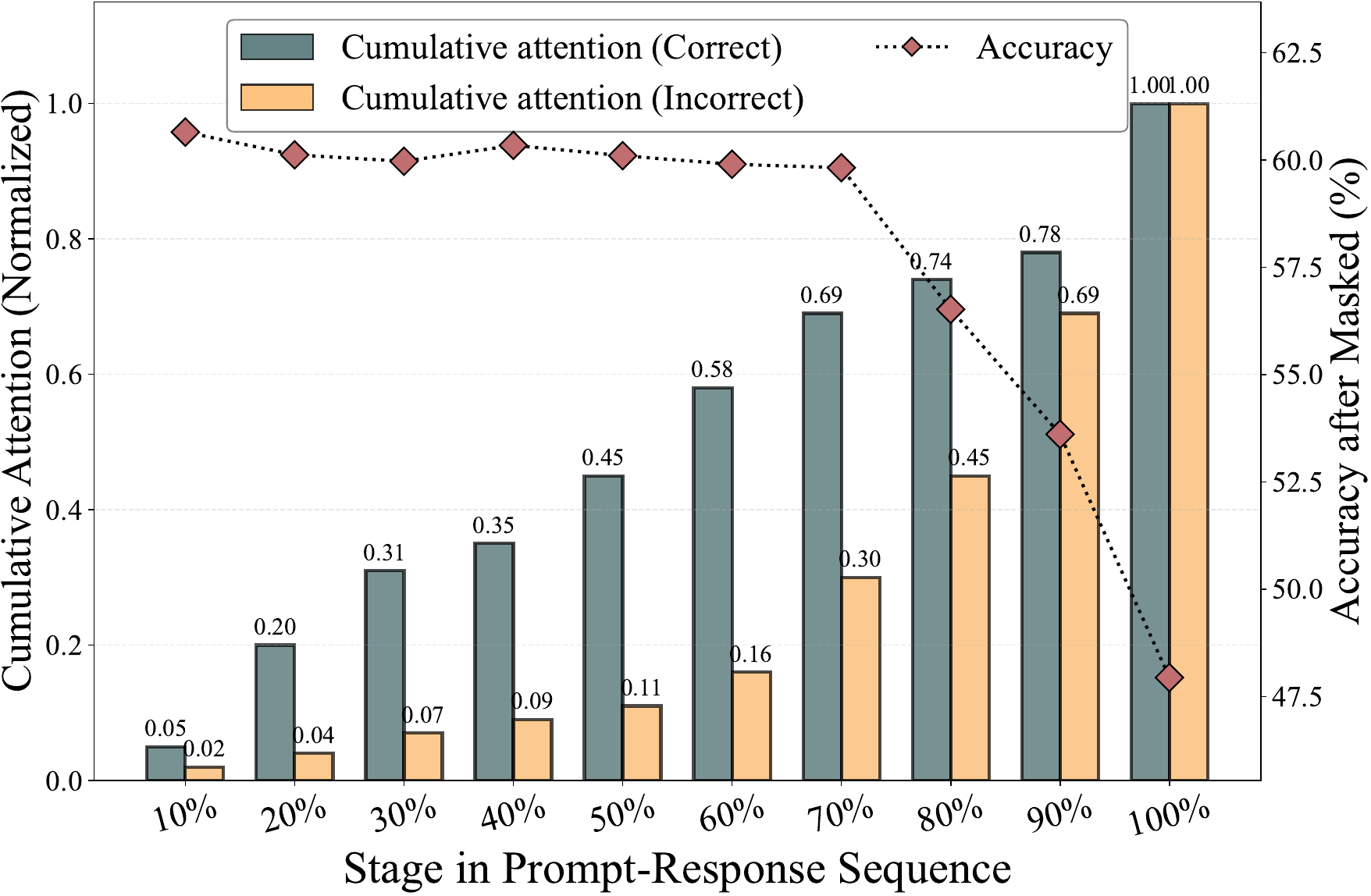}
  \caption{Attentional analysis of the reward model trained on the HH-RLHF dataset on the test set. The line graph shows the average accuracy of the reward model on the test set when masking content at different stages; the bar chart represents the normalized average cumulative attention at each stage from the beginning to the end of the joint prompt-response sequence, where “correct”, “incorrect” denote correctly classified and misclassified sequences.}
  \label{fig.attn-acc}
\end{figure}

\subsection{Discriminative Reward Modeling}

Generally, the $\theta$ parameterized discriminative reward model $\mathcal{R}_{\theta}$ consists of a decoder-only Transformer and a linear head, which takes the prompt $x$ and response $y$ as input. Formally, let the Transformer's input sequence be $s= [\langle \text{bos} \rangle ; x ; \langle \text{eos} \rangle ; y ; \langle \text{eos} \rangle]$, where $\langle \text{bos} \rangle$ and $\langle \text{eos} \rangle$ denote special tokens. Its high-dimensional representation is denoted as $\mathbf{H}^{\text{PM}}=[\mathbf{h}_1,\mathbf{h_2},...,\mathbf{h}_l]\in \mathbb{R}^{l\times d}$, where $l$ is the input sequence length and $d$ is the representation dimension. Typically, the last token's representation is projected through a linear layer to obtain the reward score $\mathcal{R}_{\theta}(x,y)$, formulated in Equation~\ref{eq.reward}.

\begin{equation}
\mathcal{R}_{\theta}(x,y) = \mathbf{W}_{\text{head}} \mathbf{h}_l\label{eq.reward}
\end{equation}

where $\mathbf{W}_{\text{head}} \in \mathbb{R}^{1\times d}$ represents the linear head parameters. The current training objective aims to amplify the reward difference between two responses under the same prompt, ensuring better responses receive higher reward values. For a given prompt $x$ along with the labeled chosen response $y_c$ and rejected response $y_r$, we require that the reward for $y_c$ should be higher than that for $y_r$, denoted as $y_c \succ y_r$. At the sample level, this objective can be achieved by minimizing the negative log-likelihood function shown in Equation~\ref{eq.loss_pm}, where $\sigma$ is the sigmoid function.

\begin{equation}
\mathcal{L}_{\text{PM}}(x,y_c,y_r) = -\log \sigma[\mathcal{R}_{\theta}(x,y_c) - \mathcal{R}_{\theta}(x,y_r)]\label{eq.loss_pm}
\end{equation}

\subsection{Perspective from Intra-Sequence Attention} \label{sec.intra}



To investigate the impact of intra-sequence attention on RM decision-making, we trained an RM on the HH-RLHF training set and conducted analysis on the test set. As shown in Figure~\ref{fig.attn-acc}, we sampled data from the test set and divided each prompt-response sequence into 10 uniform stages, analyzing the average cumulative normalized attention at each progressive stage (bar charts). The results reveal that successful decision sequences exhibit relatively steady attention growth across stages, while failed decision sequences show abrupt attention spikes in later stages. This indicates that successful decisions stem from comprehensive attention to the prompt-response relationship, whereas excessive focus on later tokens of the sequence leads to loss of premise semantics.

Furthermore, we progressively masked stages of the test sequences from front to back and calculated prediction accuracy. As shown in the line chart, the accuracy remains stable in early stages but drops sharply after 80\% masking. This demonstrates that the RM's attention generally exhibits forward-decaying characteristics. This phenomenon occurs because the reward model's backbone architecture relies entirely on Transformer decoders, where intra-sequence attention is implemented through \textbf{causal multi-head attention} (causal MHA). This unidirectional attention mechanism inherently prevents interaction with preceding tokens in the sequence, causing the reward model to disproportionately focus on later tokens of the prompt-response sequence.

\begin{figure}[t]
  \centering
  \includegraphics[width=1\linewidth]{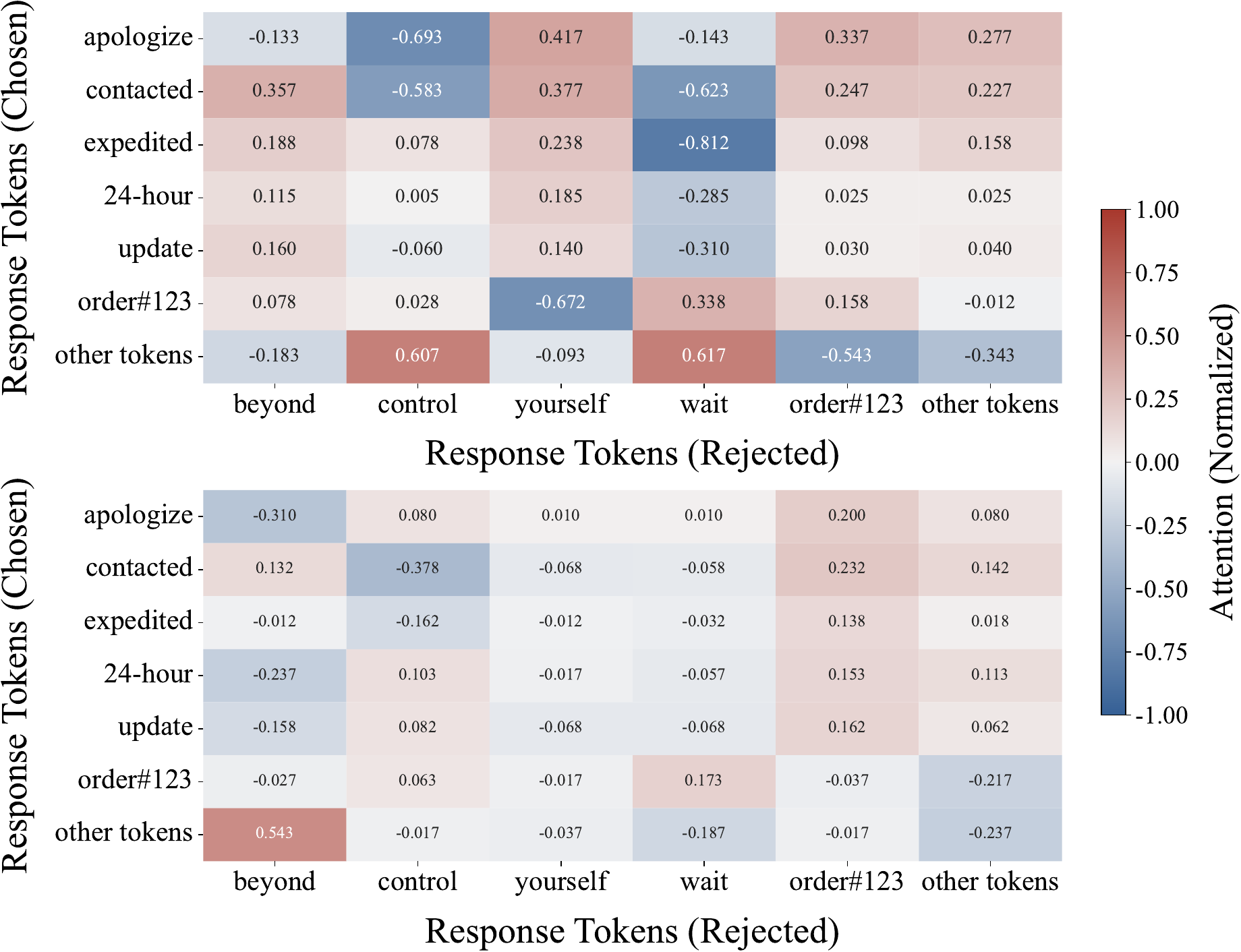}
  \caption{Attentional analysis of two reward model trained on the HH-RLHF dataset on the test set, including attention scores of tokens in the reward model encoding for correct (upper) and incorrect (bottom) decisions.}
  \label{fig.interaction-attn}
\end{figure}

\subsection{Perspective from Inter-Sequence Attention}\label{sec.inter}


To explore the impact of inter-sequence attention on RM decision-making, we trained two RMs on the HH-RLHF training set and analyzed the models' normalized attention behavior on an out-of-distribution case. We primarily focused on the following underlined key tokens, as these tokens contain critical semantic information that distinguishes the RM's judgment between the two responses.  
The \textbf{prompt} is “The order I placed yesterday \#123 hasn't arrived yet, and the logistics information has been stuck on “Shipped” status. What's going on?”, the \textbf{chosen response} is “We sincerely \underline{apologize} for any inconvenience caused! I have checked your \underline{order \#123}, and the logistics system shows a confirmed delay. I have \underline{contacted} the carrier for \underline{expedited} processing and will provide you with an \underline{update} within \underline{24 hours}. Once again, we apologize for the delay.”, and the \textbf{rejected response} is “Logistics issues are \underline{beyond} our \underline{control}. Check the app yourself for information. Order \#123 status shows ‘shipped’, just wait for delivery.”  

Figure~\ref{fig.interaction-attn} reports the normalized attention distribution (-1 to 1) of tokens for the successful and failed RMs in decision-making. It can be observed that the successful RM assigned large negative values to strongly conflicting words such as “expedited” and “wait”, as well as “apologize” and “beyond/control”, which is the key to correct decision-making. In contrast, the attention between tokens in the failed RM showed no significant differences. The \textbf{independent Siamese-encoding paradigm} in preference modeling inherently prevents early inter-sequence attention. This architectural limitation causes reward models to struggle with capturing token-level interactions, ultimately resulting in deficient global semantic perception capabilities.

\section{Method}

\begin{figure}[htbp]
  \centering
  \includegraphics[width=1\linewidth]{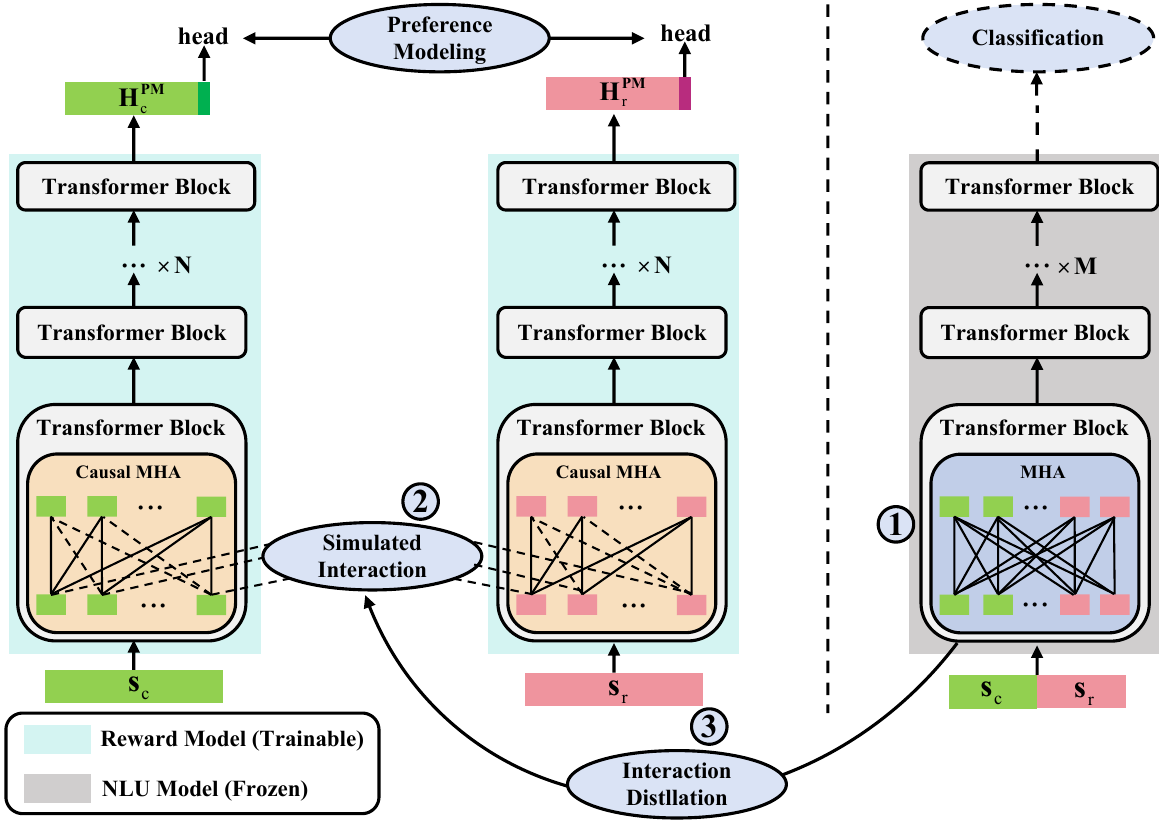}
  \caption{The interaction distillation-based preference modeling framework for reward model (\textbf{\textsc{Id-Rm}}).}
  \label{fig.framework1}
\end{figure}

Inspired by attention-level knowledge distillation~\cite{li2022virt,lu2022ernie}, we propose \emph{\textbf{“Interaction Distillation”}} to address the attention hacking and achieve more adequate preference modeling in discriminative RM, and the reward model trained with this method is termed \textbf{\textsc{Id-Rm}}. As shown in Figure~\ref{fig.framework1}, interaction distillation consists of three steps:
(1) introducing an interaction-based natural language understanding (NLU) model as a teacher to extract comprehensive attention maps across prompt-response sequences;
(2) performing identical attention computation in preference modeling as in the teacher model to simulate the teacher's interaction patterns;
(3) introducing an attentional optimization objective $\mathcal{L}_{\text{ID}}$ to align the comprehensive attention maps from the NLU model with the simulated attention maps from preference modeling, as formulated in Equation~\ref{eq.objective}. 
\begin{equation}    \min_\theta\mathbb{E}_{(x,y_c,y_r)\sim\mathcal{D}}\{\mathcal{L}_{\text{PM}}(x,y_c,y_r)+\eta\mathcal{L}_{\text{ID}}(x,y_c,y_r)\}\label{eq.objective}
\end{equation}
where $\eta$ is a hyperparameter for adjusting the weight of interaction distillation, and $\mathcal{D}$ denotes the preference dataset.



\subsection{Interaction based NLU Model as the Teacher}

\begin{figure}[htbp]
  \centering
  \includegraphics[width=0.65\linewidth]{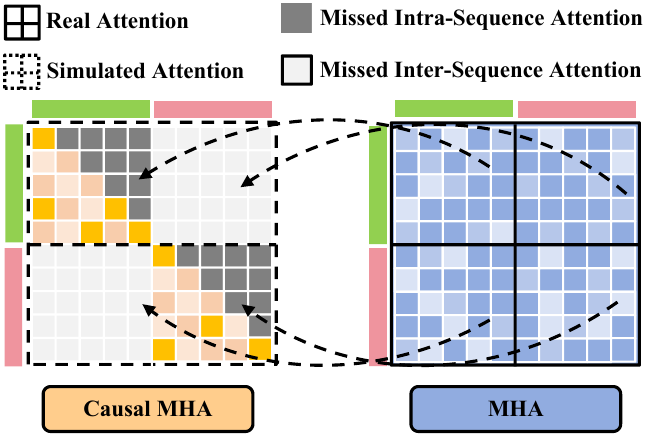}
  \caption{Attention transfer mechanism in interaction distillation}
  \label{fig.framework2}
\end{figure}

As shown in Figure~\ref{fig.framework1}, the interaction-based NLI model, composed of an encoder-only Transformer and a linear head, has emerged as the unified paradigm for natural language understanding tasks~\cite{devlin2019bert}, with the BERT family being the most prominent representative~\cite{devlin2019bert,he2022deberta}. It jointly encodes concatenated input sequences, pools representations tokens, and adapts them to classification (e.g. \{entailment, neutral, contradiction\}) via a linear layer. In our framework, we concatenate the chosen and rejected prompt-response pairs into a sequence $[s_c; s_r]$, which is mapped representation $\mathbf{H} = [\mathbf{H}_c; \mathbf{H}_r] \in \mathbb{R}^{(l_c + l_r) \times d}$. $\mathbf{H}$ is enhanced through the weighted fusion of multi-block attention maps $\mathbf{A}$ (\textbf{solid line map} in Figure~\ref{fig.framework2}) in NLI model, where $\mathbf{A} $ is formulated in Equation~\ref{eq.attn}.
\begin{equation}
    \mathbf{A}=\text{Attn}(\mathbf{Q},\mathbf{K})\overset{\rm def}{=}\text{softmax}(\frac{\mathbf{Q}\mathbf{K}^\top}{\sqrt{d}})\label{eq.attn}
\end{equation}

where $\mathbf{Q}=\mathbf{W}^{\mathbf{Q}}\mathbf{H}$, $\mathbf{K}=\mathbf{W}^{\mathbf{K}}\mathbf{H}$ denote the query and key matrices, respectively. We further decompose $\mathbf{A}$ from the perspective of $\mathbf{H}_c$ and $\mathbf{H}_r$, as formulated in Equation~\ref{eq.decompose}.



\begin{equation}
\begin{aligned}
    \mathbf{A}&=\text{Attn}([\mathbf{Q}_c;\mathbf{Q}_r],[\mathbf{K}_c;\mathbf{K}_r])
    \\&=\text{softmax}(\begin{bmatrix} \mathbf{Q}_c\mathbf{K}^\top_c/\sqrt{d} & \mathbf{Q}_c\mathbf{K}^\top_r/\sqrt{d} \\ \mathbf{Q}_r\mathbf{K}^\top_c/\sqrt{d} & \mathbf{Q}_r\mathbf{K}_r/\sqrt{d} \end{bmatrix}\\&\overset{\rm def}{=}\begin{bmatrix} \mathbf{A}_{c\rightarrow c} & \mathbf{A}_{c\rightarrow r} \\ \mathbf{A}_{r\rightarrow c} & \mathbf{A}_{r\rightarrow r} \end{bmatrix}\label{eq.decompose}
\end{aligned}
\end{equation}


As shown in Figure~\ref{fig.framework2} (\textbf{blue map}), where $\mathbf{A}_{c\rightarrow c}\in \mathbb{R}^{l_c\times l_c}$ and $\mathbf{A}_{r\rightarrow r}\in \mathbb{R}^{l_r\times l_r}$ denote the intra-sequence attention within $s_c$ or $s_r$ respectively, while $\mathbf{A}_{c\rightarrow r}\in \mathbb{R}^{l_c\times l_r}$ and $\mathbf{A}_{r\rightarrow c}\in \mathbb{R}^{l_r\times l_c}$ represent the inter-sequence attention between $s_c$ and $s_r$. The MHA enhances the model representational capacity by modeling these attention patterns. However, for reward models in preference modeling, their intra-sequence attention exhibits a “lower triangular” missing pattern, leading to the forward-decaying attention distribution over prompt-response sequences. The inter-sequence attention is completely absent due to the independent Siamese-encoding paradigm, resulting in the reward model's inability to capture token-level interactions. In conclusion, although the interaction-based NLU models with the BERT family as their backbone are significantly smaller in scale compared to the reward models based on LLM, they are indeed a more comprehensive teacher model in terms of attention.

\subsection{Simulated Interaction in Preference Modeling}


As shown in Figure~\ref{fig.framework2} (\textbf{dashed line map}), to simulate the teacher's interaction patterns, we leverage the query and key matrices to perform identical attention computation in preference modeling as in the teacher model.


The \textbf{sim}ulated interaction within preference sequences is shown in Equation~\ref{eq.pair_inter}, where $\mathbf{A}^{\text{Sim}}_{c\rightarrow c}\in \mathbb{R}^{l_c\times l_c}$ represents the bidirectional attention map of $\mathbf{H}_c^{\text{PM}}$, and $\mathbf{A}^{\text{Sim}}_{r\rightarrow r}\in \mathbb{R}^{l_r\times l_r}$ is defined similarly. These two attention maps simulate the missing bidirectional interaction in causal intra-sequence attention (\textbf{dark gray blocks}).


\begin{equation}
\begin{aligned}
    \mathbf{A}^{\text{Sim}}_{c\rightarrow c}=\text{Attn}(\mathbf{Q}^{\text{PM}}_c,\mathbf{K}^{\text{PM}}_c)\\
    \mathbf{A}^{\text{Sim}}_{r\rightarrow r}=\text{Attn}(\mathbf{Q}^{\text{PM}}_r,\mathbf{K}^{\text{PM}}_r)\label{eq.pair_inter}     
\end{aligned}   
\end{equation}

The simulated interaction between preference sequences is shown in Equation~\ref{eq.pair_interaction}, where $\mathbf{A}^{\text{Sim}}_{c\rightarrow r} \in \mathbb{R}^{l_c\times l_r}$ denotes the attention map from $\mathbf{H}_c^{\text{PM}}$ to $\mathbf{H}_r^{\text{PM}}$, and $\mathbf{A}^{\text{Sim}}_{r\rightarrow c}\in \mathbb{R}^{l_r\times l_c}$ is defined similarly. These two additional attention maps represent the missing interaction signals in absent inter-sequence attention (\textbf{light gray blocks}).


\begin{equation}
\begin{aligned}
    \mathbf{A}^{\text{Sim}}_{c\rightarrow r}=\text{Attn}(\mathbf{Q}^{\text{PM}}_c,\mathbf{K}^{\text{PM}}_r)\\
    \mathbf{A}^{\text{Sim}}_{r\rightarrow c}=\text{Attn}(\mathbf{Q}^{\text{PM}}_r,\mathbf{K}^{\text{PM}}_c)\label{eq.pair_interaction}     
\end{aligned}   
\end{equation}

\begin{table*}[t]
\footnotesize
\centering
\setlength{\tabcolsep}{6 pt}
\renewcommand{\arraystretch}{1.0} 
\begin{tabular}{l|ccc>{\columncolor{gray!15}}c|ccc>{\columncolor{gray!15}}c|ccc>{\columncolor{gray!15}}c}
\hline\hline
\multirow{2}{*}{\textbf{PPO (\textsc{Id-Rm}) vs.}} & \multicolumn{4}{c|}{\textbf{HH-RLHF (Helpful)}}                                                                                                              & \multicolumn{4}{c|}{\textbf{HH-RLHF (Harmless)}}                                                                                                             & \multicolumn{4}{c}{\textbf{TL;DR Summarization}}                                                                                                            \\
                                     & \multicolumn{1}{c}{\textbf{Win}} & \multicolumn{1}{c}{\textbf{Tie}} & \multicolumn{1}{c}{\textbf{Lose}} & \multicolumn{1}{c|}{\cellcolor{gray!15}\textbf{Win rate}} & \multicolumn{1}{c}{\textbf{Win}} & \multicolumn{1}{c}{\textbf{Tie}} & \multicolumn{1}{c}{\textbf{Lose}} & \multicolumn{1}{c|}{\cellcolor{gray!15}\textbf{Win rate}} & \multicolumn{1}{c}{\textbf{Win}} & \multicolumn{1}{c}{\textbf{Tie}} & \multicolumn{1}{c}{\textbf{Lose}} & \multicolumn{1}{c}{\cellcolor{gray!15}\textbf{Win rate}} \\\hline
SFT                                  & 39                               & 40                               & 21                                & 65.00                                & 67                               & 22                               & 11                                & 85.90                                & 89                               & 3                                & 8                                 & 91.75                                \\
DPO                                  & 40                               & 34                               & 26                                & 60.61                                & 50                               & 35                               & 15                                & 76.92                                & 83                               & 9                                & 8                                 & 91.21                                \\
PPO (BT-RM)                          & 37                               & 39                               & 24                                & 60.66                                & 42                               & 36                               & 22                                & 65.63                                & 49                               & 21                               & 30                                & 62.03                                \\
PPO (AM-RM)                          & 41                               & 32                               & 27                                & 60.29                                & 47                               & 27                               & 26                                & 64.38                                & 42                               & 29                               & 29                                & 59.15                                \\
PPO (LF-RM)                          & 43                               & 30                               & 27                                & 61.43                                & 49                               & 22                               & 29                                & 62.82                                & 51                               & 13                               & 36                                & 58.62                                \\
PPO (WARM)                           & 29                               & 50                               & 21                                & 58.00                                & 47                               & 21                               & 32                                & 59.49                                & 48                               & 12                               & 40                                & 54.55                                \\
PPO (InfoRM)                         & 25                               & 56                               & 19                                & 56.82                                & 38                               & 34                               & 28                                & 57.58                                & 39                               & 26                               & 35                                & 52.70                                \\ \hline\hline
\end{tabular}
\caption{The performance of \textsc{Id-Rm} in RLHF optimization. We report the comparison of win, tie, lose count and win rate (\%) judged by GPT-4o for \textsc{Id-Rm} against different RMs on policy models trained with PPO, SFT models, and DPO models. The reported values are averaged over 5 rounds of evaluation, with 100 samples per round.}\label{tab.rlhf}
\end{table*}

\subsection{Interaction Distillation}

To bridge the gap in token-level interactions between preference modeling and interaction-based NLU models, we propose aligning the simulated attention maps with the real attention from the interaction-based NLU model. Intuitively, the attention maps from the interaction-based model can guide preference learning toward richer interactions, as if the representations had already interacted during encoding. Through this method, we distill knowledge from the interactions and transfer it to preference modeling without introducing additional computational overhead during inference. We therefore term this mechanism interaction distillation.


Specifically, for the $M$ Transformer blocks in the reward model and the $N$ Transformer blocks in the NLU model, we define four types of simulated attention maps in each block of the reward model as $\mathbf{A}^{\text{Sim}(m)}_{c\rightarrow c}$, $\mathbf{A}^{\text{Sim}(m)}_{r\rightarrow r}$, $\mathbf{A}^{\text{Sim}(m)}_{c\rightarrow r}$, and $\mathbf{A}^{\text{Sim}(m)}_{r\rightarrow c}$, where $m \in [1,2,...,M]$. Similarly, the four types of attention maps in the NLU model are denoted as $\mathbf{A}^{(n)}_{c\rightarrow c}$, $\mathbf{A}^{(n)}_{r\rightarrow r}$, $\mathbf{A}^{(n)}_{c\rightarrow r}$, and $\mathbf{A}^{(n)}_{r\rightarrow c}$, where $n \in [1,2,...,N]$. 

Considering that the number of Transformer blocks may differ between the reward model and the NLU model, we select the attention maps from the top $K$ blocks (By default, $K=1$) and align them using the $L_2$ norm in the loss function~\ref{eq.align}. This loss function incorporates sub-objectives for both intra-sequence and inter-sequence distillation.




\vspace{-13pt}
\begin{equation}
\begin{aligned}
    &\mathcal{L}_{\text{ID}}(x,y_c,y_r) = \frac{1}{4K}\sum_{(m,n)=(M,N)}^{(M-K,N-K)}\\&(\underbrace{\frac{1}{l_c}\|\mathbf{A}^{(n)}_{c\rightarrow c}-\mathbf{A}^{\text{Sim}(m)}_{c\rightarrow c}\|_2+\frac{1}{l_r}\|\mathbf{A}^{(n)}_{r\rightarrow r}-\mathbf{A}^{\text{Sim}(m)}_{r\rightarrow r}\|_2}_{\text{Intra-sequence distillation}}
    \\&+\underbrace{\frac{1}{l_c}\|\mathbf{A}^{(n)}_{c\rightarrow r}-\mathbf{A}^{\text{Sim}(m)}_{c\rightarrow r}\|_2+\frac{1}{l_r
    }\|\mathbf{A}^{(n)}_{r\rightarrow c}-\mathbf{A}^{\text{Sim}(m)}_{r\rightarrow c}\|_2}_{\text{Inter-sequence distillation}} )\label{eq.align}    
\end{aligned}
\end{equation}

This objective can be seamlessly integrated with reward models without altering the RM's architecture or inference paradigm, thus introducing no additional inference overhead. During training, it only requires forward propagation through the teacher model to obtain attention maps, resulting in merely a 5\% average increase in time per batch and a 3\% growth in GPU memory usage.

\begin{table*}[t]
\footnotesize
\centering
\setlength{\tabcolsep}{6 pt}
\renewcommand{\arraystretch}{1.0} 
\begin{tabular}{l|cc|cccc|c}
\hline\hline
\multirow{2}{*}{\textbf{RM}} & \multicolumn{2}{c|}{\textbf{RMB}}     & \multicolumn{4}{c|}{\textbf{Reward Bench}}                                  & \multirow{2}{*}{\textbf{Avg.}} \\
                             & \textbf{Harmless} & \textbf{Helpful} & \textbf{Chat}  & \textbf{Chat-Hard} & \textbf{Safety} & \textbf{Reasoning} &                                \\\hline
DPO                          & 57.87            & 53.44             & \textbf{92.84} & 49.67              & 73.87           & 57.67              & 64.22                          \\
BT-RM                        & 56.63            & 56.71             & 90.03          & 49.99              & 74.21           & 61.42              & 64.83                          \\
AM-RM                        & 57.27            & 55.13             & 91.24          & 50.92              & 74.43           & 62.75              & 65.29                          \\
LF-RM                        & 57.78            & 56.38             & 92.45          & 51.21              & 75.99           & 62.19              & 66.00                          \\
WARM                         & 56.84            & 57.58             & 91.02          & 52.98              & 77.42  & 63.44              & 66.83                          \\
InfoRM                       & 58.90            & 56.02             & 91.45          & 53.12              & \textbf{79.12}  & 62.90              & 66.92                          \\
\rowcolor{gray!20}\textbf{\textsc{Id-Rm}}               & \textbf{59.07}   & \textbf{57.11}    & 92.23          & \textbf{54.56}     & 78.34           & \textbf{67.23}     & \textbf{68.08}      \\ \hline\hline          
\end{tabular}
\caption{Accuracy (\%) of \textsc{Id-Rm} in OOD preference perception. Bold values indicate the best result in each group.}\label{tab.ood}
\end{table*}

\section{Experiments}

\subsection{Experimental Setup}





In this study, we employed LLaMA3~\cite{grattafiori2024llama} with 8 billion parameters as the backbone for the di s reward model, and adopted an interaction-based NLU model based on DeBERTa-large~\cite{he2022deberta} as the teacher model. The teacher model was fine-tuned on natural language understanding dataset SNLI~\cite{bowman2015large}. To evaluate stability and generalizability of the reward signals provided by \textsc{Id-Rm}, we conducted a series of experiments in both reinforcement learning from human feedback (RLHF) and out-of-distribution (OOD) preference perception. The between-group differences in the experimental results reported in this study were all statistically significant at the 5\% confidence level, as verified by \textbf{Wilcoxon signed-rank tests}.~\footnote{Detailed setup, implementations, and case studies provided in the supplementary materials.}

\subsubsection{RLHF Setting.} In RLHF tasks, we use \textsc{Id-Rm} to provide supervision signals for the policy model and optimize the policy model using \textbf{Proximal Policy Optimization (PPO)}~\cite{schulman2017proximal}. The policy model is initialized via supervised fine-tuning (SFT) beforehand. For the \textbf{general dialogue task}, we used the HH-RLHF dataset~\cite{bai2022training} for PPO optimization, focusing on the helpful and harmless perspectives. For the \textbf{summarization} task, we employed the TL;DR dataset~\cite{volske2017tl}.


For evaluation, we used the policy model trained with \textsc{Id-Rm} to answer questions from the test set, while the SFT model and the policy model trained via PPO with Bradley-Terry RM served as opponents answering the same test questions. Existing research demonstrates strong correlation between GPT-4o's response assessments and human evaluation results~\cite{zheng2023judging,chang2024survey}. Thus, we evaluated their responses using GPT-4o~\cite{achiam2023gpt} and calculated the \textbf{win rate} of \textsc{Id-Rm}, where the win rate is calculated as the proportion of winning samples to the total number of winning and losing samples.

\subsubsection{OOD Preference Perception Setting.} In OOD preference perception, to explore whether \textsc{Id-Rm} can provide generalizable signals in out-of-distribution tasks, we performed inference-time verification on the world reward model benchmark RMB~\cite{zhou2024rmb} and Reward Bench~\cite{lambert2024rewardbench}, using accuracy for evaluation. Among them, RMB includes two dimensions: helpfulness and harmlessness, while Reward Bench comprises four subtasks: Chat, Chat-Hard, Safety, and Reasoning.


\subsubsection{Baselines.} We employ SFT model and Direct Preference Optimization (DPO) model~\cite{rafailov2023direct} as baselines for RLHF tasks. Additionally, we select SOTA reward model optimization methods as baselines: Bradley-Terry-based BT-RM~\cite{schulman2017proximal}, adaptive-margin based AM-RM~\cite{touvron2023llama}, label-flipping-based LF-RM~\cite{burns2024weak,wang2024reward}, weight averaging based WARM~\cite{rame2024warm}, and information-bottleneck-based InfoRM~\cite{miao2024inform}, which primarily address noise handling in the data.

\begin{table*}[t]
\footnotesize
\centering
\setlength{\tabcolsep}{5.5 pt}
\renewcommand{\arraystretch}{1.0} 
\begin{tabular}{l|c|ccc|ccc}\hline\hline
\multicolumn{1}{c|}{\multirow{2}{*}{\textbf{\textsc{Id-Rm}}}} & \multirow{2}{*}{\textbf{Study}} & \multicolumn{3}{c|}{\textbf{HH-RLHF (Win rate)}}                                                                                      & \multicolumn{3}{c}{\textbf{RMB (OOD Accuracy)}}                                                                                      \\
\multicolumn{1}{c|}{}                                         &                                 & \multicolumn{1}{l}{\textbf{Helpful}} & \multicolumn{1}{c}{\textbf{Harmless}} & \multicolumn{1}{c|}{\textbf{$\Delta$ (Avg.)}} & \multicolumn{1}{c}{\textbf{Harmless}} & \multicolumn{1}{c}{\textbf{Helpful}} & \multicolumn{1}{c}{\textbf{$\Delta$ (Avg.)}} \\ \hline
\rowcolor{gray!20}
Teacher: DeBERTa w/ tuning on SNLI (\textbf{base})             &                                 & 60.66                                & 65.63                                 & -                                             & 59.07                                & 57.11                                 & -                                             \\
Teacher: DeBERTa w/o tuning on SNLI                            & \multirow{5}{*}{Ablation}       & 54.14                                & 61.33                                 & \cellcolor{blue!10}-5.41                     & 56.24                                & 56.83                                 & \cellcolor{blue!25}-2.06                     \\
Teacher: BERT w/ tuning on SNLI                                &                                 & 60.74                                & 65.23                                 & -0.16                                         & 58.20                                & 58.77                                 & -0.11                                         \\
Teacher: BERT w/o tuning on SNLI                               &                                 & 54.92                                & 62.09                                 & -4.64                                         & 56.55                                & 57.21                                 & \cellcolor{blue!10}-1.71                     \\
Loss: w/o Intra-sequence distillation                               &                                 & 52.45                                & 57.23                                 & \cellcolor{blue!25}-8.30                     & 57.04                                & 57.19                                 & -1.48                                         \\
Loss: w/o Inter-sequence distillation                               &                                 & 58.44                                & 62.98                                 & -2.44                                         & 58.22                                & 57.86                                 & -0.55                                         \\ \hline
w/ Adaptive margining                                                  & \multirow{3}{*}{Compatibility}  & 61.89                                & 66.79                                 & \cellcolor{red!10}1.20                       & 61.22                                & 60.03                                 & 2.04                      \\
w/ Label flipping                                                   &                                 & 62.07                                & 66.21                                 & 1.00                       & 62.19                                & 59.44                                 & \cellcolor{red!10}2.22                      \\ 
w/ Weight averaging                                 &                                 & 63.03                                & 66.52                                 & \cellcolor{red!25}1.63                                     & 62.01                                & 60.02                                 & \cellcolor{red!25}2.43       \\                             
\hline\hline
\end{tabular}
\caption{Ablation and compatibility study results of \textsc{Id-Rm}. We report the win rate (\%) of \textsc{Id-Rm} against BT-RM on HH-RLHF and the OOD accuracy (\%) on RMB. $\Delta$ (Avg.) indicates the average change relative to the base configuration.}\label{tab.ablation}
\end{table*}

\subsection{Main Results}

\subsubsection{RLHF tasks}


We trained the policy model using PPO with \textsc{Id-RM}. This policy model serves to compete against SFT and DPO models, as well as other PPO-trained policy models using the 4 baseline reward models.

Table~\ref{tab.rlhf} demonstrates that \textsc{Id-Rm} \textbf{achieved win rates exceeding 0.5 against all baselines}. Notably, it exhibited the \textbf{strongest advantage in harmlessness scenarios}. This is because harmlessness assessment requires not only contextual comprehension but also heightened sensitivity to critical tokens in prompt-response sequences. Through interaction distillation, \textsc{Id-Rm} enriches both intra-sequence and inter-sequence attention in preference modeling, thereby achieving more comprehensive semantic perception.

\subsubsection{OOD Preference Perception}

Firstly, we train reward models on the HH-RLHF dataset and evaluate them on both OOD reward model datasets and more comprehensive reward modeling benchmarks. Table~\ref{tab.ood} reports that \textsc{Id-Rm} achieves \textbf{superior performance compared to other reward modeling methods} on RMB's two tasks and Reward Bench's Chat-Hard and Reasoning tasks. With average accuracy of 68.25\%, the results demonstrate that \textsc{Id-Rm}, by modeling more comprehensive attention mechanisms, not only adjusts attention allocation within sequences but also sensitively captures token-level interactions between preference pairs. This enables the learning of more generalizable information and improves accuracy in recognizing out-of-distribution preference pairs.

\subsection{Ablation \& Compatibility Study}

To investigate the necessity of each component in interaction distillation and its compatibility with other reward modeling optimization algorithms, we conducted ablation studies and compatibility studies. Table~\ref{tab.ablation} reports the relative average win rate against BT-RM on the HH-RLHF task and the average accuracy on the RMB benchmark under different settings.  

\subsubsection{Ablation Study.} 



We conducted ablation studies from two perspectives: the teacher model and the loss function.

Regarding the \textbf{teacher model}, compared to DeBERTa fine-tuned on the SNLI dataset, we found that \textbf{replacing the backbone with BERT did not lead to a significant performance drop} in \textsc{Id-Rm}. However, using a \textbf{teacher model that was not fine-tuned on SNLI resulted in severe performance degradation}. This is because the primary pretraining task for the BERT family is masked token prediction, and further fine-tuning on the SNLI dataset enables comprehensive modeling of interactions both within and between different sequences. This lays a solid foundation for the teacher model to effectively distill the full attention map into preference modeling.

Regarding the \textbf{loss function}, we observed that both intra-sequence and inter-sequence distillation significantly impact the objectives of interaction distillation. Notably, \textbf{removing intra-sequence distillation led to a more severe performance decline} compared to inter-sequence distillation. This indicates that in preference modeling, attending to the prompt within a sequence is more important than focusing on token-level interactions between different sequences.

\subsubsection{Compatibility Study}


We employ three easily integrable algorithms: adaptive margining, label flipping, and weight averaging~\cite{burns2024weak,touvron2023llama,wang2024reward,rame2024warm} for compatibility experiments. As shown in Table~\ref{tab.ablation}, the RM's performance \textbf{achieves further improvement when integrating these algorithms, with weight averaging yielding the most significant boost}. This demonstrates that interactive distillation effectively maintains compatibility with other optimization algorithms without altering the original model's architecture.

\subsection{Hyperparameter Sensitivity Study}

\begin{figure}[t]
  \centering
  \includegraphics[width=1\linewidth]{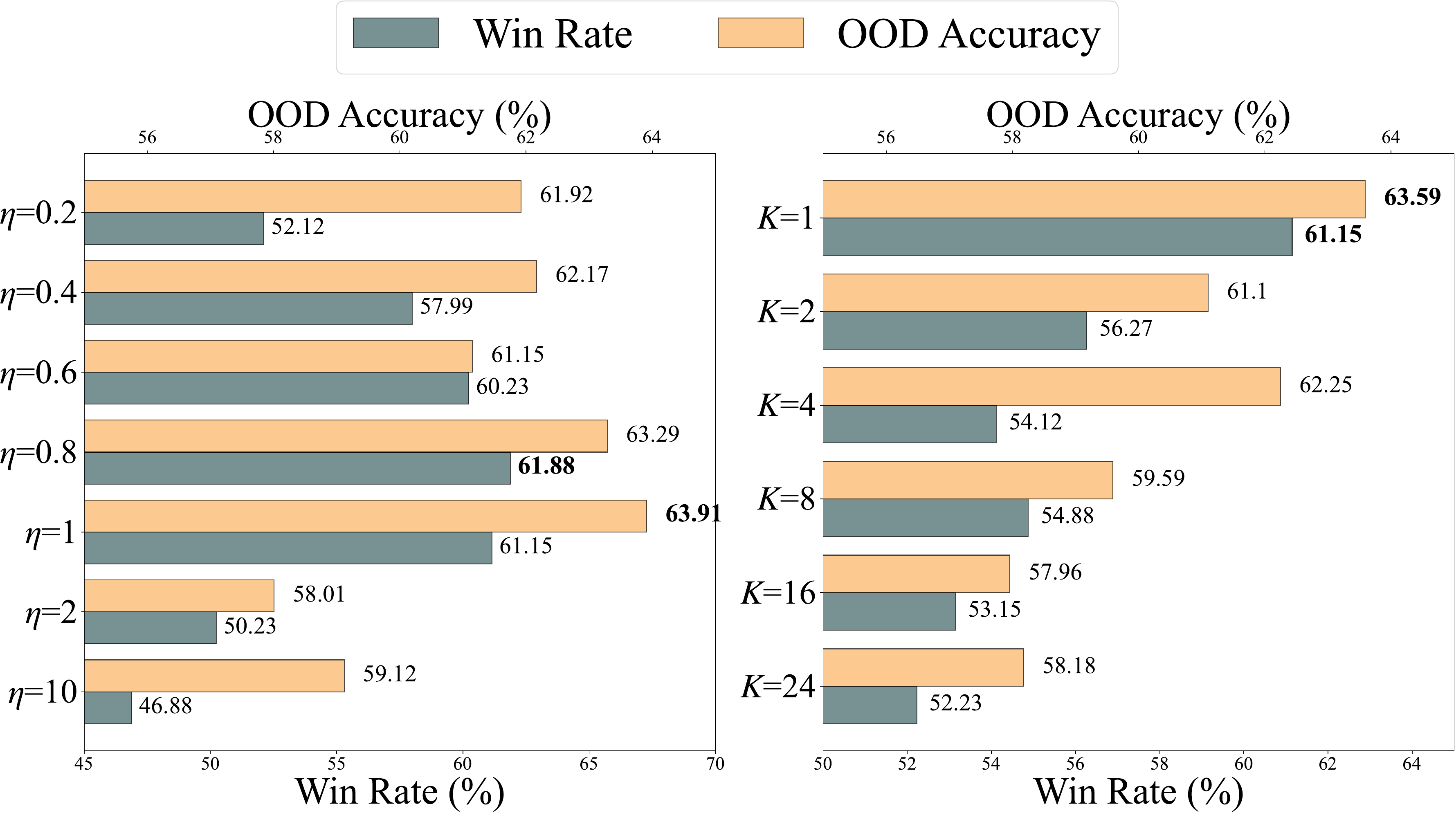}
  \caption{Hyperparameter sensitivity study, for $\eta$ (left) and $K$ (right) respectively.}
  \label{fig.sen}
\end{figure}

To investigate two crucial hyperparameters in interaction distillation, namely $\eta$ (the weight of interaction distillation) and $K$ (the number of supervised Transformer blocks in interaction distillation), we conduct controlled experiments on each hyperparameter separately. Figure~\ref{fig.sen} reports the results on the same evaluation benchmark as the ablation study.

\subsubsection{Hyperparameter $\eta$}


We experiment with $\eta \in \{0.2, 0.4, 0.6, 0.8, 1, 2, 10\}$. When $\eta$ is set to 0.8 and 1, \textsc{Id-RM} achieves the highest OOD accuracy and RLHF win rate respectively, indicating that the importance of interaction distillation is comparable to that of preference modeling. Moreover, the performance of \textsc{Id-RM} remains stable when $\eta$ falls within the range of 0.4--1.  

\subsubsection{Hyperparameter $K$}


We experiment with $K \in \{1,2,4,8,16,24\}$. When $K$ is 1 (i.e., interaction distillation is performed on the last layer of the supervised reward model), the highest out-of-distribution average accuracy and RLHF win rate are achieved. This is because the number of Transformer block of the reward model and the teacher model is inconsistent, making the information processing mechanisms at the same block non-equivalent. Therefore, we only need to supervise the attention states of the last block.

\section{Related Work}

\subsection{Reward Modeling in RLHF}
Reinforcement learning from human feedback ~\cite{ouyang2022training,bai2022training} provides a crucial and straightforward approach to align large language models with human preferences, constituting an essential training procedure for current mainstream LLMs~\cite{achiam2023gpt,touvron2023llama,yang2024qwen2,guo2025deepseek} and improving model utility, honesty, and harmlessness~\cite{ethayarajh2022understanding}. 

As the core component of RLHF, the reward model provides critical feedback for responses sampled by the policy model~\cite{dou2024multi,wang2024secrets,dou2025alleviating}. The paradigm of reward modeling can be divided into generative reward models~\cite{mahan2024generative,zhang2024generative,liu2025inference} and discriminative reward models. However, discriminative reward hacking~\cite{miao2024inform,fu2025reward} poses a critical challenge for current reward models, as reward models are often hacked by biases in the data, leading to erroneous judgments. While existing optimization approaches primarily focus on addressing noise in the data~\cite{touvron2023llama,burns2024weak,rame2024warm,miao2024inform}, they largely overlook the inadequate attention mechanisms in the preference modeling framework. We pioneer the investigation of attention hacking stemming from structural defects in reward models.


\subsection{Knowledge Distillation for Language Models}



The knowledge distillation techniques for large language models have evolved from simple output imitation to a series of sophisticated methodologies. Among them, a foundational and widely applied technique is supervised fine-tuning (SFT), which involves training the student model using prompt-response data pairs generated by the teacher model~\cite{taori2023stanford}. The success of such methods largely depends on the quality and diversity of the training data. To this end, the research community has explored various data generation strategies: extensions of self-instruct~\cite{wang2022selfinstruct}, which bootstrap generation from a small seed set of instructions, while more advanced approaches like Evol-Instruct evolve initial instructions by increasing their complexity and depth to enhance the model's reasoning capabilities~\cite{xu2023wizardlm}. 

For white-box models where internal parameters are accessible, feature-based distillation methods aim to minimize the differences between the teacher and student models in terms of output probability distributions~\cite{gu2024minillm}. Furthermore, a cutting-edge direction is self-improvement, where models iteratively refine and enhance their knowledge through frameworks like self-play~\cite{chen2024selfplay} or self-rewarding mechanisms~\cite{yuan2024selfrewarding}. However, this study is the first to propose interaction distillation specifically for LLM based reward models, achieving knowledge distillation at the attention level.

\section{Conclusion}


We propose a novel solution termed “Interaction Distillation” to address the “attention hacking” in discriminative reward models. This approach introduces an interaction-based NLU model to distill its comprehensive attention patterns into reward models. Remarkably, it effectively rectifies the model's inadequate token-level interaction via an attentional distillation objective, without introducing inference costs. Compared to reward modeling optimization methods that primarily address data noise, our approach generates more stable and generalizable reward signals. This suggests we can move beyond merely mitigating reward model bias at the data noise level, instead capture the intrinsic structure of complex human preferences through attention mechanisms.


\appendix

\bibliographystyle{named}
\bibliography{aaai24,ijcai,nml,attn}

\end{document}